(This is a draft verion of the manuscript that was accepted in IEEE BESC 2020 conference)

# Social Network Analysis of Hadith Narrators from Sahih Bukhari

Tanvir Alam*, Jens Schneider
College of Scicence and Egnigeering,
Hamad Bin Khalifa University, Doha,
Qatar
*Correspondence: talam@hbku.edu.qa

***Abstract*—The ahadith (plural of hadith), prophetic traditions for the Muslims around the world, are narrations originating from the sayings and the deeds of Prophet Muhammad (pbuh). They are considered one of the fundamental sources of Islamic legislation along with the Quran. The list of persons involved in the narration of each hadith is carefully scrutinized by scholars studying the hadith, with respect to their reputation and authenticity of the hadith. This is due to the ahadith's legislative importance in Islamic principles. There were many narrators who contributed to this responsibility of preserving prophetic narrations over the centuries. But to date, no systematic and comprehensive study, based on the social network, has been adapted to understand the contribution of early hadith narrators and the propagation of hadith across generations. In this study, we represented the chain of narrators of the hadith collection from Sahih Bukhari as a social graph. Based on social network analysis (SNA) on this graph, we found that the network of narrators is a scale-free network. We identified a list of influential narrators from the companions (e.g., Abu Hurairah, Ibn Abbas, Ibn Umar, etc.) as well as the narrators from the second and third-generation (e.g., Shu'bah bin al-Hajjaj, Az-Zuhri, Sufyan bin 'Uyaynah, Sufyan bin Sa'id Ath-Thawri, etc.) who contribute significantly in the propagation of hadith collected in Sahih Bukhari. We discovered sixteen communities from the narrators of Sahih Bukhari. In each of these communities, there are other narrators who contributed significantly to the propagation of prophetic narrations. We also found that most narrators were centered in Makkah and Madinah (in today's Saudi Arabia) in the era of companions and, then, gradually the center of hadith narrators shifted towards Kufa, Baghdad (in today's Iraq) and central Asia (e.g., Uzbekistan, Turkmenistan) over a period of time. To the best of our knowledge, this the first comprehensive and systematic study based on SNA, representing the narrators as a social graph to analyze their contribution to the preservation and propagation of hadith.**

## I. INTRODUCTION

The ahadith (أحاديث), plural of hadith (حديث), the recorded words, actions, and tacit approval from the Prophet Muhammad (peace be upon him (phuh)), are one of the most important sources of legislation, along with the Quran, for the Muslims. Throughout the history of an early Muslim scholar, there had been a tremendous effort to preserve the Hadiths verbatim [1]. Each hadeeth consists of two parts. The first part, sanad (سند), consists of a chain of narrators who transmitted the saying of the Prophet Muhammad (phuh). It begins with the last transmitter who collected it in his Hadith collection book and ends with the companion who narrated it from the Prophet Muhammad (pbuh). The second part is called matan (المتن), describing the saying, action, physical description, tacit approval of the Prophet Muhammad (phuh). The chain of narrators (sanad, سند) mentioned at the beginning of each Hadith serves as a testament to its strength and authenticity.

During the lifetime of Muhammad (pbuh), the Prophet of Muslims, there was no urgency in the Muslim community to write down the saying or actions of the prophet, even though some of his companions started writing some of His sayings and actions. But after the death of Prophet Muhammad (pbuh) the importance of preserving His words become prominent. This stage of hadith preservation by the companions of Prophet Muhammad (pbuh) started at the mid of the first century AH ("*Anno Hegirae*", starting with the migration to Medina in 622 CE), when Prophet Muhammad (pbuh) passed away. Among the leading companions who narrated the most ahadith are Abu Hurayrah, Abdullah Ibn Abbas, Abdullah Ibn Amr Ibn Al-Aas, Abu Bakr, Ibn Umar, etc. [1]. After Islam spread into North Africa, India, and Europe, the preservation of ahadith become instrumental in spreading the true message of Islam. This was the era of Tabioon ("*followers*" or "*successors*") (1st century AH) and at this stage, caliph Umar ibn Abdul Aziz (reign 99 AH – 101 AH) also ordered the leading scholars to compile and preserve the ahadith. In this era, Sad ibn Ibrahim and Ibn Shihab Az-Zuhri were also requested to preserve the ahadith. Imam Ibn Shihab Az-Zuhri was the first compiler to record the biographies of narrators. After the era of Tahioon, the era of Tabi-Tabioon (students of Tabioon), mainly in the 2nd century AH, begins. In this period, the ahadith were preserved systematically in written books. One of the most famous and the earliest books was Al-Muwatta by Malik Bin Anas. Other prominent scholars in this era who contributed significantly in this era were Al-Awzaee (from Syria), Abdullah ibn Al-Mubarak (from Basra), Hamad ibn Salamah (from Basra), Sufyan ath-Thawree (from Kufa). After the first three generations (companions, Tabioon, Tabi-Tabioon) of hadith collection, formal research on the collected ahadith begins. In this era of Tabi-Tabi-Tabioon (3rd century AH) some leading scholars of hadith conducted research on the ahadith collection and grouped some of them as authentic ahadith. From this era, the most famous six books (al-kutub as-sittah, اَلْكُتُب اَلسِتَّة) that shaped the Islamic law and legislation are: Sahih Bukhari, Sahih Muslim, Sunan Abu Dawood, Jami At-Tirmidhi, Sunan As-Sughra (by an-Nasaee), and either Ibn-Majah or Muwatta Malik.

A lot of sincere efforts have been made by the early generation of Muslim scholars to preserve the authenticity of the prophetic text and actions [2]. This network of hadith narrators based on the tradition of hadith from one generation to the next makes hadith collections some of the most authentic and well-preserved religious texts for the Muslim community. In addition to the holy Quran, hadith books are considered as the source of religious knowledge and verdict for Islamic jurisprudence [3]. Consequently, there is a need for a systematic approach to analyze the contribution of scholars in the preservation and propagation of ahadith across generations. There exist many studies pertinent to the classification and mining of the matan (text) of the Hadith leveraging data mining and natural language processing techniques [4]. But no comprehensive study has been made, so far, considering the full chain of narrators for hadiths to evaluate the contribution and influence of hadith narrators in a systematic fashion.

To fill this gap, we, in this study, represented the chain of narrators from Sahih Bukhari as a graph and analyzed the underlying network of this graph. We, then, used social network analysis (SNA) techniques to generate insights from this graph. To the best of our knowledge, this is the first attempt to apply SNA in a systematic fashion to understand the contribution of hadith narrators in the preservation and propagation of prophetic narrations.

## II. Results

### A. *Hadith Narrators in Sahih Bukhari*

First, From Sahih Bukhari, we found a total of 7,370 hadiths narrated by a total of 1,372 narrators. From this collection of ahadith, we found a single chain of narrators for the majority of the ahadith. For only 153 ahadith (2.07% of total 7,370 hadith), we found multiple chains. Note that for this study we do not yet consider merging "duplicate" ahadith that differ mostly in their chain. This is left for future work in collaboration with Islamic scholars. However, we provide our data as electronic appendix to facilitate such research.

### B. *Era and Locality of the Haidth Narrators in Sahih Bukhari*

From the first generation, era of companions (Sahabah), we found 195 narrators. From the second generation, era of successors (Tabioon) we found 588 narrators. From the era of Tabi-Tabioon and Tabi-Tabi-Tabioon we found 326 and 263 narrators respectively for the hadiths mentioned in Sahih Bukhari. Table 1 highlight the cities and the number of narrators for each generation.

### C. *Centrality of the Haidth Narrators in Sahih Bukhari*

We calculated the degree centrality of each narrator in the network (see more details on degree centrality in the "Materials and Methods" section). Briefly, if narrator A describes a hadith heard from Prophet Muhammad (phuh) to the narrator B; and narrator B propagates it to narrator C, then we represented these three narrators (A, B and C) as a directed graph with three nodes. A propagated the hadith to B, so the outdegree of narrator A is zero. As B heard the hadith from A and propagated to C, then for B the indegree (received from A) was one and outdegree (propagated to C) was one. C only received the hadith from B, so the indegree of C was one. In this way, we calculated the indegree and outdegree of each narrator for each hadith. From Sahih Bukhari, we found that most of the narrators have low degree (indegree and outdegree), and few of them have high degree centrality (Figure 1(a)). Among the narrators Abu Hurairah, Anas bin Malik, Hisham bin 'Urwa, ibn Abbas and Shu'bah bin al-Hajjaj, ibn Umar, Aisha bint Abi Bakr, Jabir ibn 'Abdullah, Abu Sa'id al-Khudri , Az-Zuhri were among the top-ranked ten narrators with 83, 76, 55, 55, 52, 51, 48, 46, 44, 43 outdegree respectively. This means these narrators were involved in preserving the highest number of ahadith in Sahih Bukhari (Figure 1 (b)).

TABLE I. Shifting the Center of Hadith Transmission over Time.

| Country | City | First era | Second era | Third era | Fourth era |
|---|---|---|---|---|---|
| Afghanistan | Bulkh | | | 1 | 6 |
| Egypt | Egypt[1] | 1 | 9 | 8 | 8 |
| Ethiopia | Abyssinia | 3 | 2 | | |
| Iran | al-Ray | | 6 | 1 | 1 |
| Iran | Hiran | | | 4 | 4 |
| Iran | Nisapur | | | 1 | 9 |
| Iran | Khurasan | | 1 | 7 | 5 |
| Iran | Hamadan | | 17 | 5 | 3 |
| Iraq | al-Wasit | | 3 | 8 | 13 |
| Iraq | Baghdad | | | 7 | 33 |
| Iraq | Kufa | 3 | 125 | 67 | 35 |
| Iraq | Basra | 5 | 105 | 106 | 61 |
| Saudi Arabia | Al-Ta'if | | 4 | 1 | 1 |
| Saudi Arabia | Al-Yamama | | 1 | 4 | 1 |
| Saudi Arabia | Hijaz | 14 | 4 | | |
| Saudi Arabia | Makkah | 70 | 39 | 5 | 2 |
| Saudi Arabia | Madinah | 87 | 202 | 38 | 11 |
| Syria | al-Hims | | 6 | 9 | 5 |
| Syria | Damascus | | 7 | 9 | 7 |
| Turkmenistan | Merv | | 1 | 4 | 18 |
| Uzbekistan | Bukhara | | | | 5 |
| Yemen | Al-Azdi | | 1 | | 4 |

City centers having at least five narrators are highlighted in this table. Color code highlights the relative frequency of narrators for each era in different cities (red, yellow and green color scales represent the high, moderate and few numbers of narrators respectively). [1]: This may refer to the city Fustat in today's Cairo.

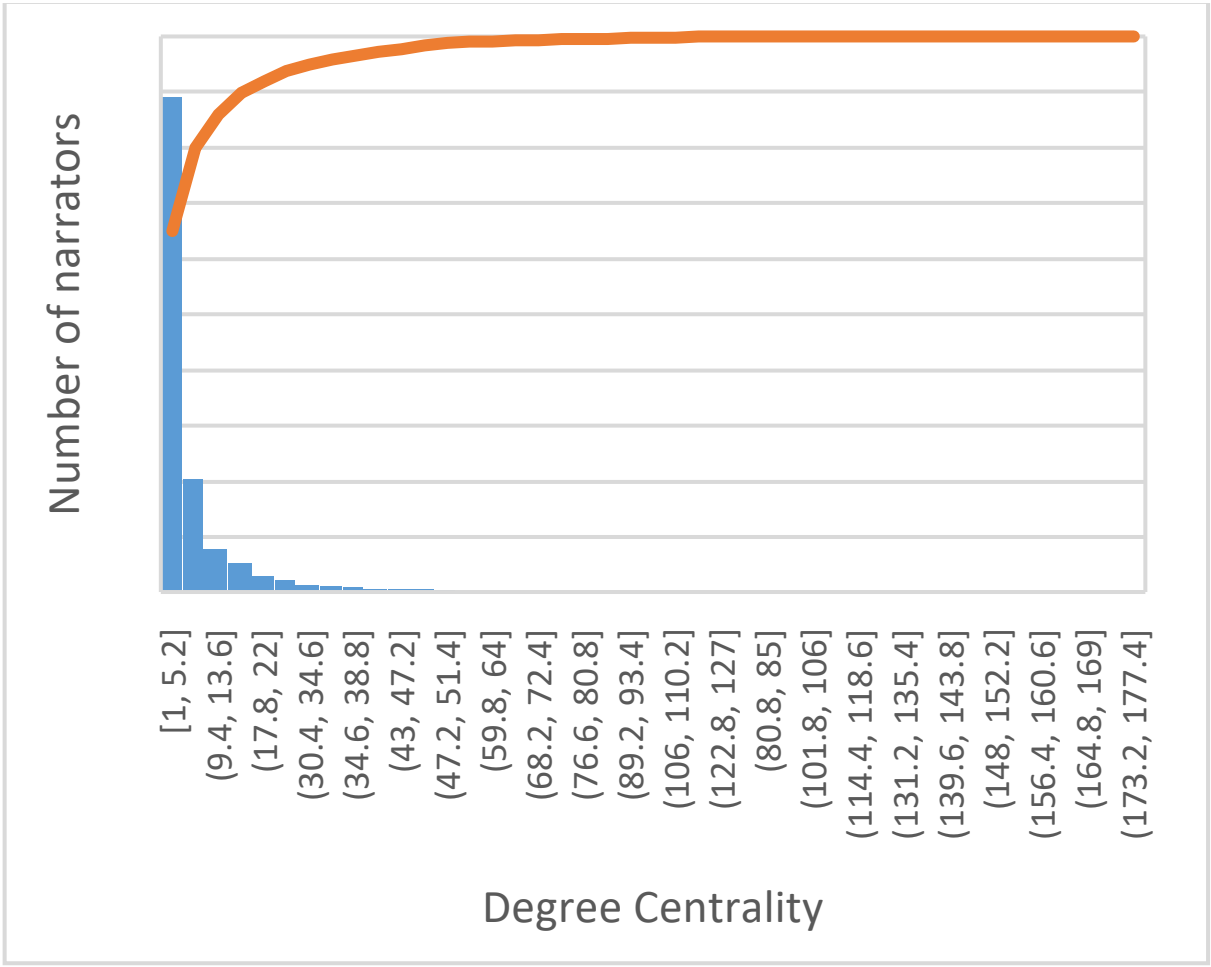

(a)

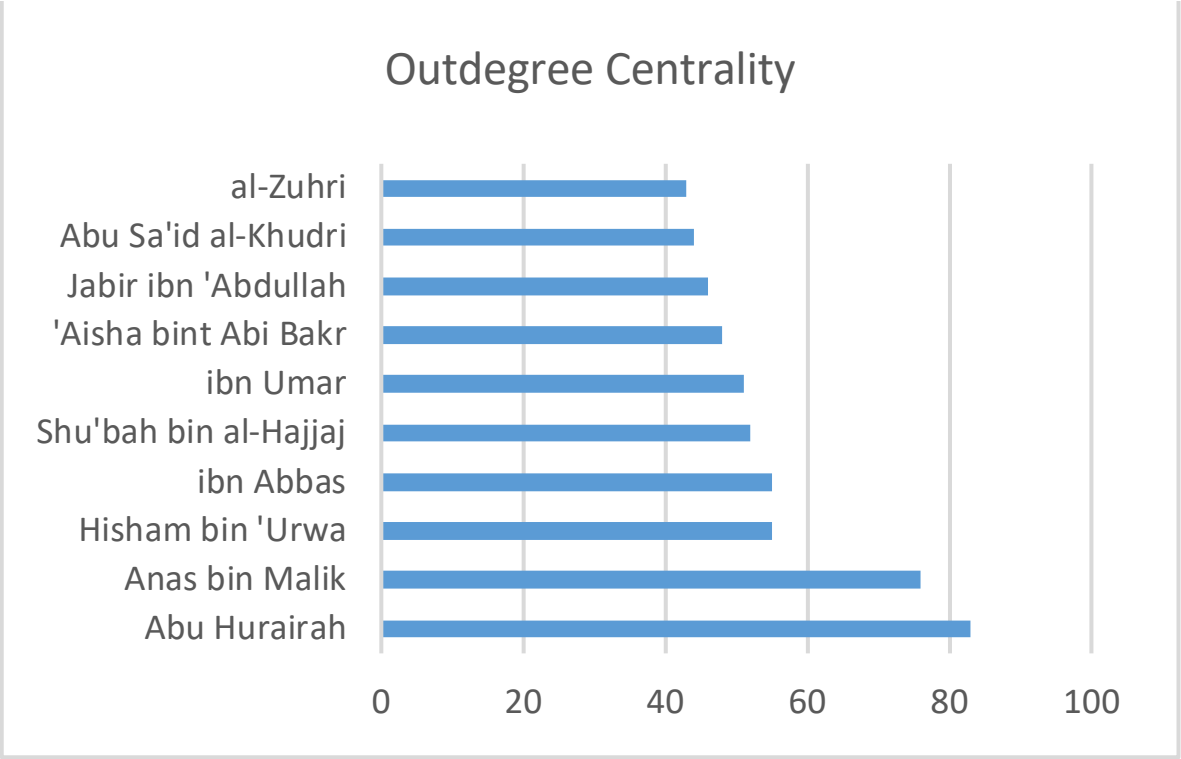

*(b)*

**Fig 1.** (a ) Distribution of Degree centrality for all narrators in Sahih Bukhari. (b) Top ten narrators with higest number (outdegree) of hadith narratives.

Based on the PageRank algorithm [5] on the directed graph that we developed based on the narrators of Sahih Bukhari (see more details on centrality in the "Materials and Methods" section)) , we found that Shu'bah bin al-Hajjaj, who was born in Basra, Iraq, was the most important node in the network of Sahih Bukhari (Figure 2). In Figure 2 (a), we highlighted the most influential narrators of the Sahih Bukhari based on the PageRank algorithm. Az-Zuhri, born in Madinah, Saudi Arabia, was the second most important narrator. Sufyan bin 'Uyaynah, Sufyan bin Sa'id Ath-Thawri and Al-Fadl bin Dakayn Abu Na'eem were the third, fourth and fifth most important narrators from Sahih Bukhari according to their PageRank. Figure 2 (b) shows he distribution of indegree and outdegree for the top-ranked narrators from Sahih Bukhari based on PageRank.

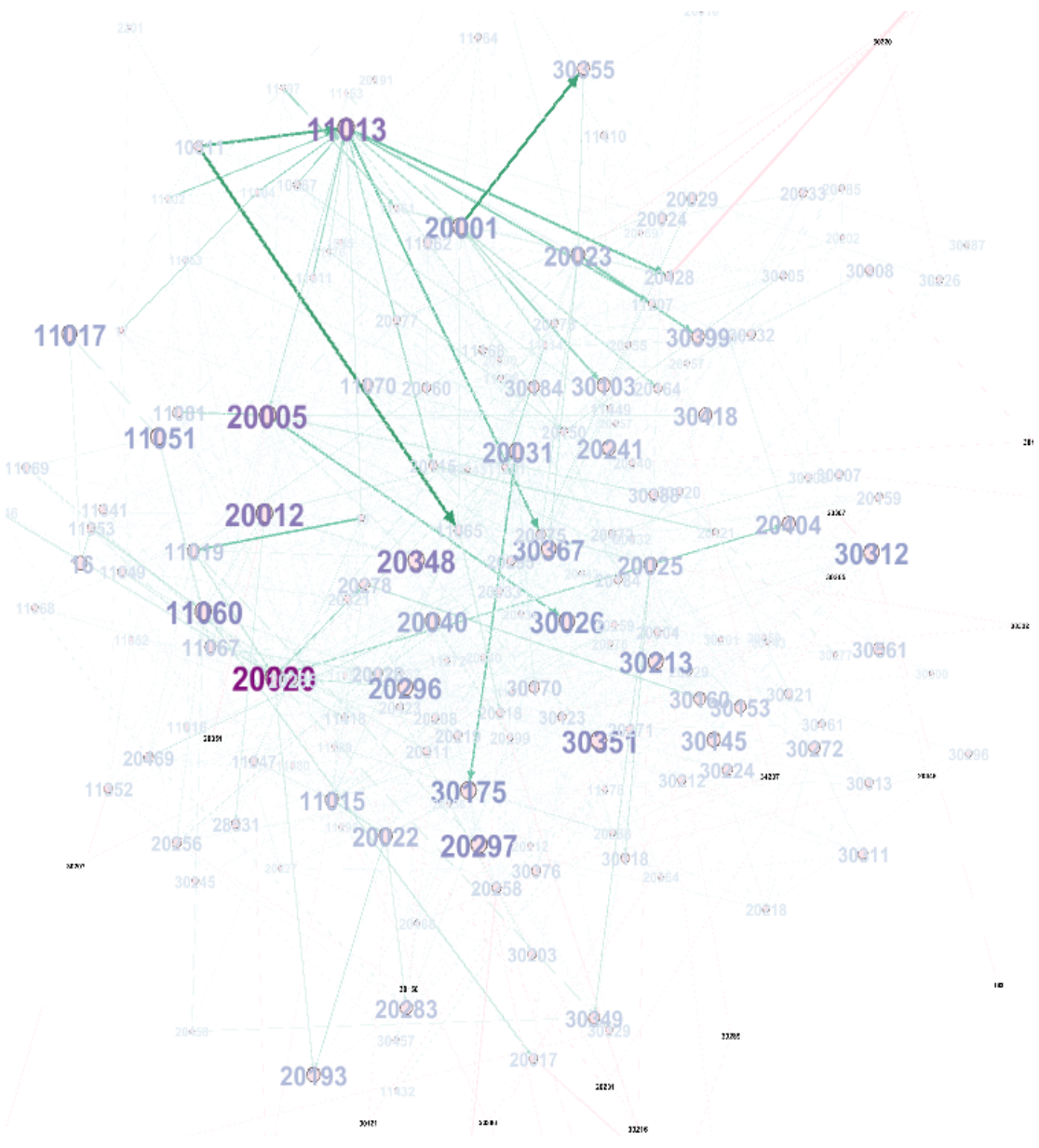

(a)

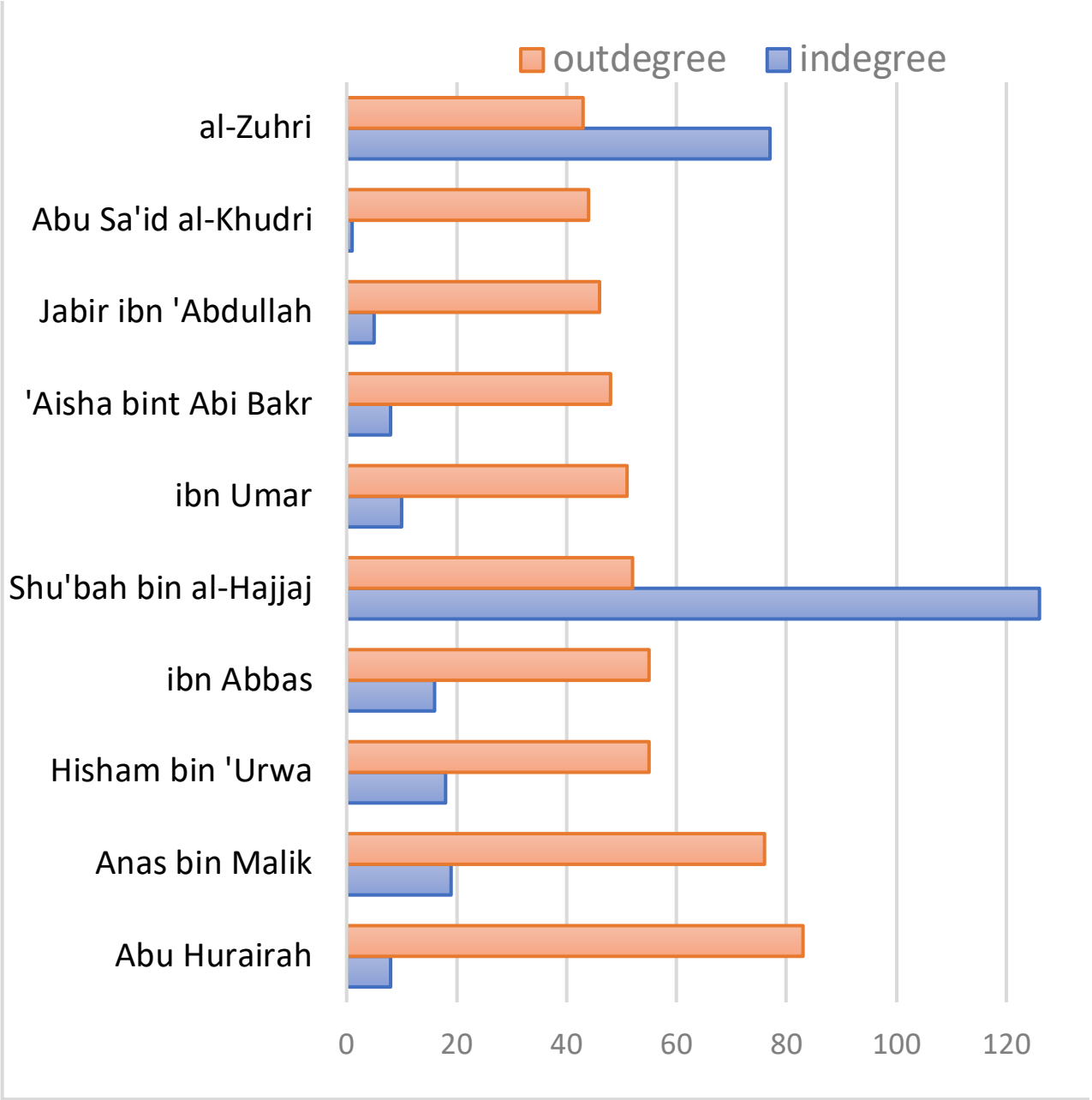

(b)

**Fig 2:** (a) PageRank -based graph for the narrators from Sahih Bukhari. Each number in the graph represents a narrator ID from http://muslimscholars.info/. Label size and color of each node is proportional to the PageRank value (higher the value more important the node is involved for the propagation of ahadith) of each node. 20020 : Shu'bah bin al-Hajjaj; 11013: Az-Zuhri; 20005: Sufyan bin 'Uyaynah; 20012: Sufyan bin Sa'id Ath-Thawr; 20348: al-Fadl bin Dakayn Abu Na'eem ; 11060: Sulaiman al-A'mash ; 20297: Hmam bin Yahya bin Dinar; 30351: 'Abdullah bin Muhammad al-Musandi; 30175: Masdad bin Masrhad ; 30026: 'Ali bin al-Madini; 30312: Ishaq bin Shahyn al-Wasti; 20001: Imam Malik; Graph was generated using Gephi. (b) Indegree and outdegree for the top ten narrators from Sahih Bukhari based on PageRank

We also applied the betweenness algorithm [6] to calculate the importance of a node (narrator) in social network of narrators in terms of the number of shortest paths (between other nodes) on which a node exists to propagate the hadith in the chain of narrations. We found that Sufyan bin 'Uyaynah, who was born in Kufa, Iraq, had the highest betweenness centrality. Figure 3 highlights the top-ranked ten narrators in terms of betweenness.

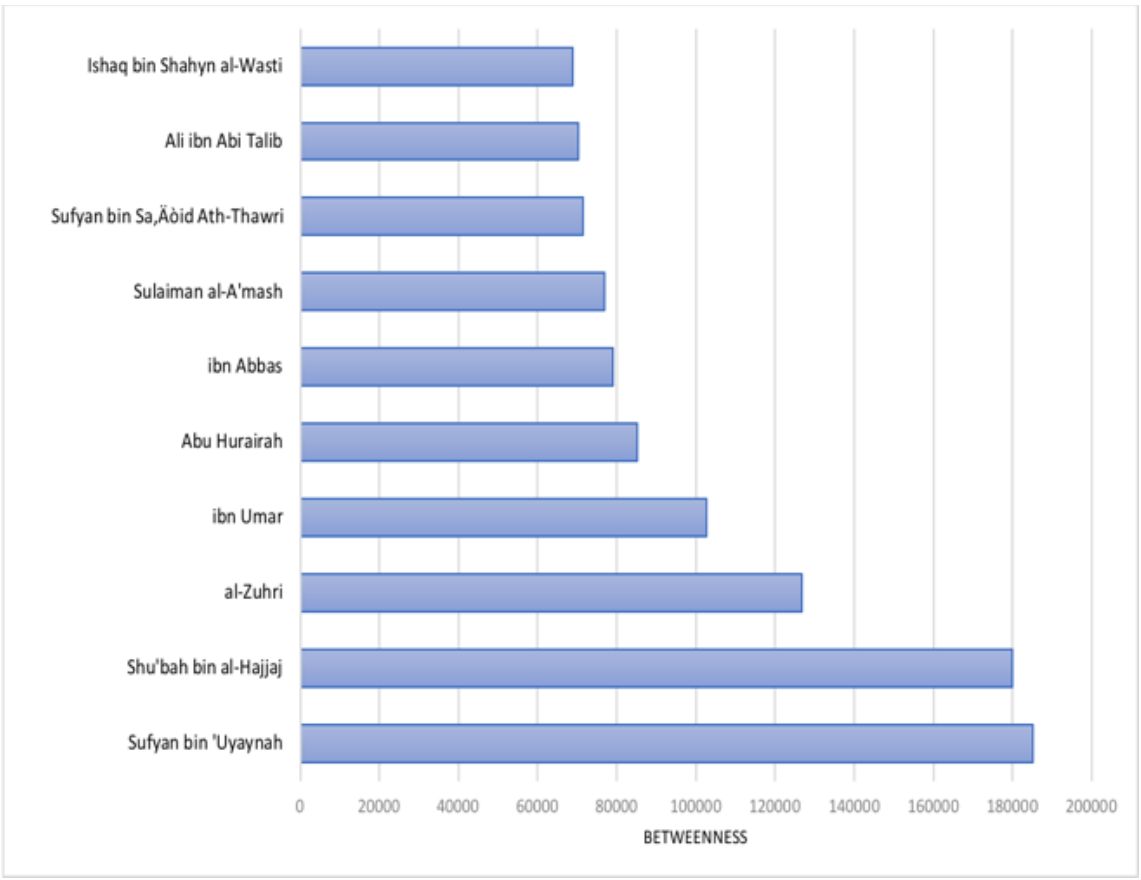

Fig 3: Top listed narrators for their contribution based on betweenness centrality

We found that Shu'bah bin al-Hajjaj , Az-Zuhri, Ibn Umar, Abu Hurairah, Ibn Abbas, Sulaiman al-A'mash,

Sufyan bin Saod Ath-Thawri, Ali ibn Abi Talib, Ishaq bin Shahyn al-Wasti are among the top ten ranked narrators based on betweenness centrality.

### *D. Community Detection among the Haidth Narrators in Sahih Bukhari*

We found total 16 communities of narrators from the list of all narrators of Bukhari (Figure 4 (a)). To visualize the whole network based on modularity, we applied a circle packing layout (https://datavizcatalogue.com/methods/circle_packing.html) with the hierarchical order of modularity and betweenness to generate Figure 4 (b), which highlights the community of narrators in the network of narrators of hadith from Sahih Bukhari.

## III. DISCUSSINOS

In this study, we leveraged the potential of social network graphs as a digital representation of the narrators of Sahih Bukhari. This study is unique when compared to previous works on the narrators of Sahih Bukhari along some important lines. This is the first study where chains of narrators are represented as a social network graph, which enabled us to measure the contribution of the narrators, using different centrality measures, for preserving this holy text in a quantitative manner. We discovered that the network of narrators from the Sahih Bukhari are a type of “small world networks” [7] having high density of small local clusters of narrators and sparse in other parts of network (Figure 1). Figure 1 highlighted that most of the narrators had small degree (they are part of small number of narrations) and few narrators had high degree (they propagated a large number of hadith). We also showed that, in the era of companions, Makkah and Madinah (both in today’s Saudi Arabia) were the centers of hadith narrations for Sahih Bukhari (Table 1). Then gradually over time (second and third era), this center shifted towards Kufa and Basra (in today’s Iraq), and in the fourth era it further expands into several regions like Bukhara (currently in Uzbekistan), Merv (currently in Turkmenistan), etc. (Table 1).

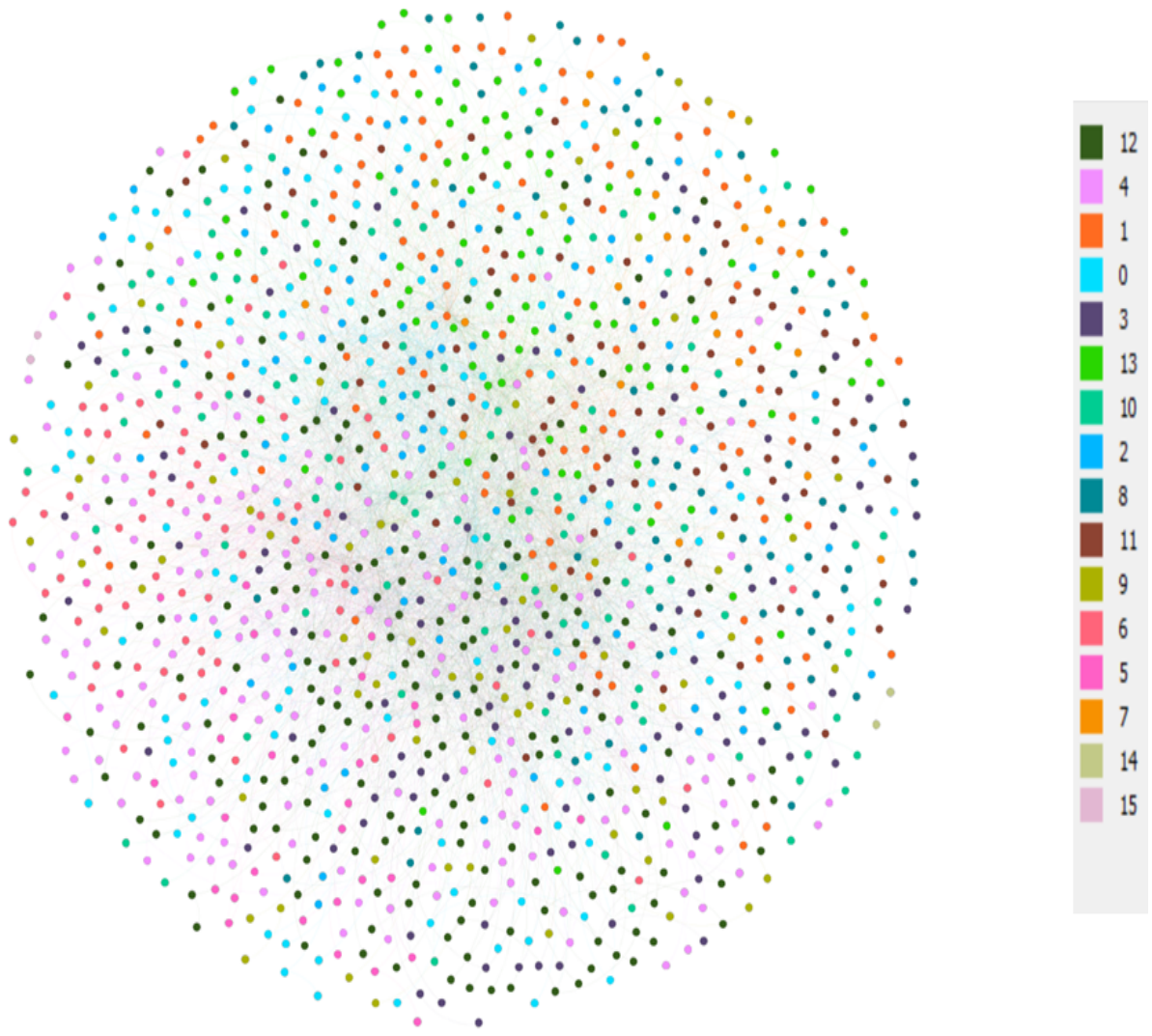

(a)

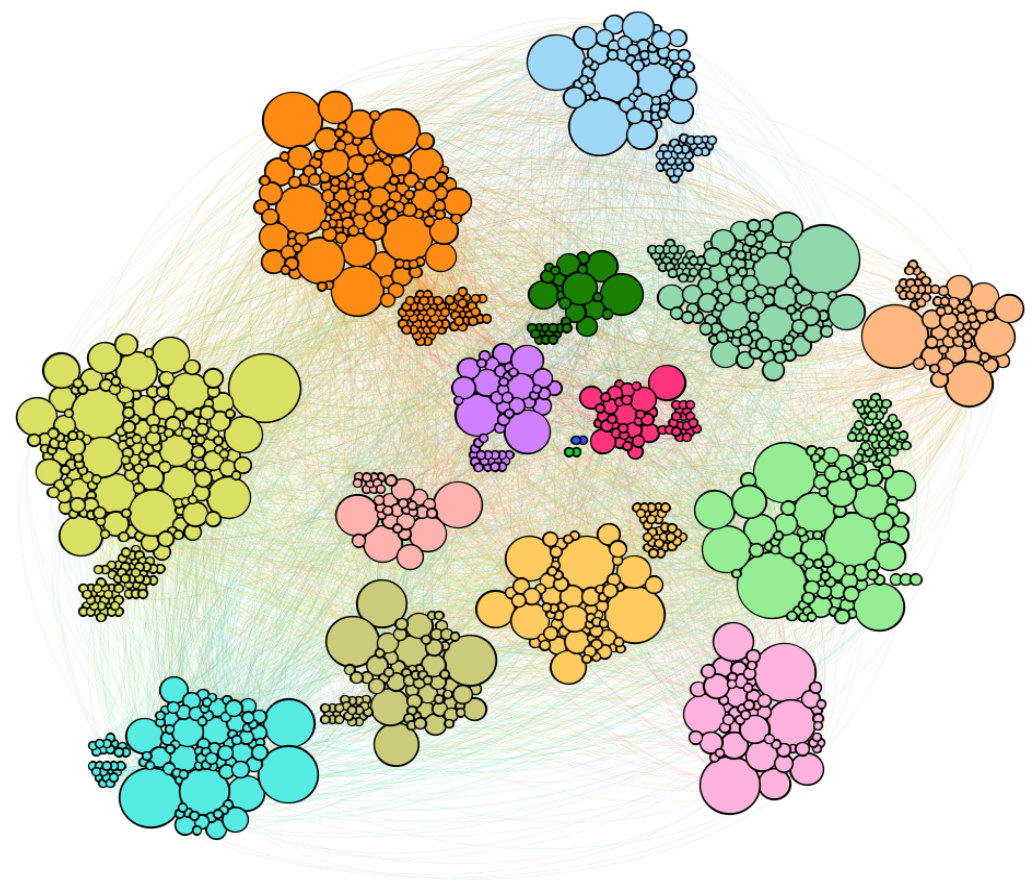

(b)

**Figure 4.** (a) Community of hadith narrators from Sahih Bukhari. Nodes are colored based on the community. The layout of the graph was generated based on Fruchterman-Reingold algorithm [8]. (b) Visualization of community using circle pack layout. Each node is group based on community. The size of node represents the betweenness centrality of node.

Based on outdegree (number of hadith propagated to other narrators), we found that most of the narrators with high outdegree are from the first era (Figure 1(b)). But when we ranked the narrators based on PageRank, which depends upon both indegree (heard from the Prophet Muhammad (pbuh) or other narrators) and outdegree (propagated to other narrators), we found that most of the top-ranked narrators are from the second or third era (Figure 2(b)). This is quite conceivable as the narrators from the second and third era received many narrations from the narrators of the first era (high indegree compared to outdegree). Consequently, PageRank identifies the narrators from the second and third era as most influential narrators.

Interestingly Ibn Umar, Ibn Abbas and Abu Hurairah, the prominent companions of the Prophet (phuh), all were ranked in the top ten influential narrators considering both outdegree and betweenness centrality. This indicates that companions of the Prophet (phuh), from the first era, made a significant contribution in the preservation and propagation of ahadith. Following the footsteps of the companions of the Prophet (pbuh), scholars from the next era continued their dedication to propagate the texts of Sahih Bukhari. This can be conceived from our result where the narrators (e.g., Az-Zuhri , Sufyan bin Said Ath-Thawri, Sufyan bin 'Uyaynah, Shu'bah bin al-Hajjaj) from that second or third era were also ranked among the top ten influential narrators based on different centrality measurements (betweenness, degree, PageRank) (Figure 1 (b), Figure 2 (b), Figure 3). We thus conclude that not only the scholars from the first era but the scholars from the next eras contributed significantly to propagate and preserve the text of Sahih Bukhari. For each narrator from Sahih Bukhari, Supplementary Table S1 provided the details of all SNA attributes that is calculated in this study.

We also discovered 16 different communities among the narrators based on the number of narrations preserved through their contribution. Each of these communities needs to be investigated in more detail to understand the narrations and the locality of these narrators. To facilitate such research, we provide our data as an electronic appendix (Supplementary Table S2). Communities are an important field of study in hadith sciences since they allow scholars to establish the veracity of the chain of traditions. Each connection in the chain is composed of a teacher (narrator) and a student (listener). To be accepted, the general requirement is that teacher and student have met face to face. This, however, can be hard to establish. If either teacher or student are of immaculate reputation and explicitly mention the meeting, this can be accepted as evidence. However, such cases are far from covering the entire canon of ahadith. Therefore, scholars also accept if, across multiple ahadith, a narrator can be linked to a certain place and time through multiple chains (typically more than four as a threshold). The rationale is that travelling scholars exchanged their knowledge at centers of learning, and such a contact would have resulted in multiple evidence in the form of adjacency in our social graph. Studying such communities (and their connections) in the future could therefore allow to propagate measures of strength and veracity through the social graph or help understand the connection between communities through time and space. In addition to such community analysis, we will also enhance our study for other prominent hadith books such as Sahih Muslim, Sunan Abu Dawood, Jami at-Tirmidhi, Sunan As-Sughra (by An-Nasaee), Sunan ibn Majah and Muwatta Maliki in future.

## IV. Materials and Methods

### A. *Data Sources and Affiliations*

We collected ahadith mentioned in Sahih Bukhari from www.sunnah.com. We also collected the full chain of narrations of hadiths from www.qaalarasulallah.com . All the data sources are publicly available, and we collected the data from above sources using custom scripts written in Python and Java..

### B. *Network representation of the chain of narrators*

We represented the narrators of each hadith as a node in graph. If narrator A conveys a hadith to narrator B, then we consider there are two nodes (A and B) and a directed edge indicating a "narrated to" relation: from A (source/teacher) to B (target/student). The weight of an edge was represented as the number of different hadith that was conveyed from narrator A to narrator B. The result is an edge-weighted directed acyclic graph.

From the collection of Sahih Bukhari, we found many hadith mentioning the words of Prophet with single chain of narrators [9]. Here is an example of hadith with single chain of narration:

حَدَّثَنَا الْحُمَيْدِيُّ عَبْدُ اللَّهِ بْنُ الزُّبَيْرِ، قَالَ حَدَّثَنَا سُفْيَانُ، قَالَ حَدَّثَنَا يَحْيَى" بْنُ سَعِيدٍ الأَنْصَارِيُّ، قَالَ أَخْبَرَنِي مُحَمَّدُ بْنُ إِبْرَاهِيمَ التَّيْمِيُّ، أَنَّهُ سَمِعَ عَلْقَمَةَ بْنَ وَقَّاصٍ اللَّيْثِيَّ، يَقُولُ سَمِعْتُ عُمَرَ بْنَ الْخَطَّابِ ـ رضى الله عنه ـ عَلَى الْمِنْبَرِ، قَالَ سَمِعْتُ رَسُولَ اللَّهِ صلى الله عليه وسلم يَقُولُ " إِنَّمَا الأَعْمَالُ بِالنِّيَّاتِ، وَإِنَّمَا لِكُلِّ امْرِئٍ مَا نَوَى، فَمَنْ كَانَتْ هِجْرَتُهُ إِلَى دُنْيَا يُصِيبُهَا أَوْ إِلَى امْرَأَةٍ يَنْكِحُهَا فَهِجْرَتُهُ إِلَى مَا هَاجَرَ إِلَيْهِ "

"I heard Allah's Apostle saying, "The reward of deeds depends upon the intentions and every person will get the reward according to what he has intended. So, whoever emigrated for worldly benefits or for a woman to marry, his emigration was for what he emigrated for" (Book of Revelation, Hadith no: 1)". The list of narrators for this hadith was Abdullah bin al-Zubair bin 'Isa →Sufyan bin 'Uyaynah → Yahya bin Sa'id al-Ansari → Muhammad bin Ibrahim bin al-Harith → Alqama bin Waqqas → Umar ibn al-Khattab. Figure 5 shows the representation of the narrators for this particular hadith in a directed graph.

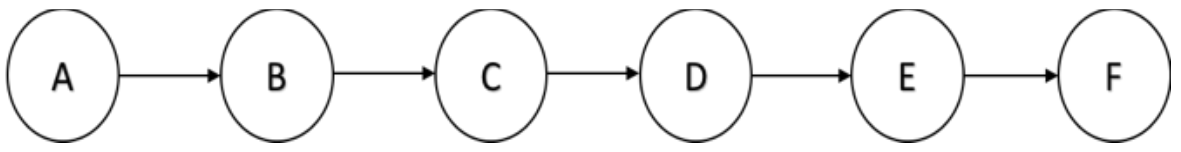

Fig 5. Directed graph representation for the narrators in a hadith; A: Abdullah bin al-Zubair bin 'Isa; B: Sufyan bin 'Uyaynah ; C: Yahya bin Sa'id al-Ansari; D: Muhammad bin Ibrahim bin al-Harith; E: Alqama bin Waqqas; F: Umar ibn al-Khattab

Not all the hadith are related to the words of Prophet (pbuh), rather some of them involved the deeds or tacit approval from the Prophet that was mentioned by the narrators. Here is an example of such hadith with single chain of narration:

حَدَّثَنَا مُطَرِّفٌ أَبُو مُصْعَبٍ، قَالَ حَدَّثَنَا عَبْدُ الرَّحْمَنِ بْنُ أَبِي الْمَوَالِي، عَنْ" مُحَمَّدِ بْنِ الْمُنْكَدِرِ، قَالَ رَأَيْتُ جَابِرَ بْنَ عَبْدِ اللَّهِ يُصَلِّي فِي ثَوْبٍ وَاحِدٍ وَقَالَ رَأَيْتُ النَّبِيَّ صلى الله عليه وسلم يُصَلِّي فِي ثَوْبٍ."

"I saw Jabir bin `Abdullah praying in a single garment and he said that he had seen the Prophet praying in a single garment. (Book of Prayer, Hadith no: 5)". The list of narrators for this hadith is Matraf bin 'Abdullah → Abdur Rahman bin Abi al-Mawal → Muhammad bin al-Munkdar bin 'Abdullah

In some cases, the chain of narrations become more complex as it may contain multiple chain of narrators [9]. The multiple chains of narrations were manually curated to generate the graph. Here is an example of hadith with multiple chain:

حَدَّثَنَا أَبُو الْوَلِيدِ، حَدَّثَنَا هَمَّامٌ، عَنْ قَتَادَةَ، عَنْ أَنَسٍ، حَدَّثَنَا مُحَمَّدُ بْنُ" سِنَانٍ، حَدَّثَنَا هَمَّامٌ، عَنْ قَتَادَةَ، عَنْ أَنَسٍ ـ رضى الله عنه ـ أَنَّ عَبْدَ، الرَّحْمَنِ بْنَ عَوْفٍ وَالزُّبَيْرَ شَكَوَا إِلَى النَّبِيِّ صلى الله عليه وسلم ـ يَعْنِي الْقَمْلَ ـ فَأَرْخَصَ لَهُمَا فِي الْحَرِيرِ، فَرَأَيْتُهُ عَلَيْهِمَا فِي غَزَاةٍ."

"Abdur Rahman bin Auf and Az-Zubair complained to the Prophet, i.e. about the lice (that caused itching) so he allowed them to wear silken clothes. I saw them wearing such clothes in a holy battle." (Book of Jihad, Hadith no: 133)". Here, we have two chain of narrators that eventually merge: (a) Hisham bin 'Abdul Malik al-Tayalasi → Hmam bin Yahya bin Dinar → Qatada → Anas bin Malik, (b) Muhammad bin Snan bin Yazid → Hmam bin Yahya bin Dinar → Qatada → Anas bin Malik.

### C. *Brief Biography of the Narrators*

Scholars of hadith, in the early stages of the development of the hadith sciences, grouped the narrators into various group. One of the famous categories was suggested by Ibn Hajar al-'Asqlaanee, in his famous book entitled Taqreeb at Tahdheeb [10], where he classified the narrator into 12 generations based on their lifespan [3]. The first era was for the companions (generation 0), the second era was for Tabioon (generation 1-6), the third era was for Tabi-Tabioon (generation 7-9) and the fourth era was Tabi-Tabi-Tabioon (generation 10-12). We collected the brief details (name, location, lifespan) of narrators from www.muslimscholars.info. We also considered this generations and grouped them into four eras based on the lifespan of narrators as mentioned in the same source.

### D. Social Network Analysis

Once we built the whole network of narrators from the collection of Sahih Bukhari, we applied different social network analysis techniques on this graph to identify the important nodes (narrators) that were influential in authentic propagation of hadith from generation to generation. We applied several centrality measurements (degree centrality, PageRank, betweenness centrality etc. [11, 12] to identify the influential nodes, in this network of narrators.

We applied degree centrality [13] to calculate the number of narrations that were propagated through a narrator for the preservation of prophetic texts. To measure the influence a node that act as a broker between communities we also applied betweenness centrality [14] for each narrator of this network. PageRank, a variant of Eigenvalue centrality and WeightedDegree is based on the degree (indegree and outdegree) of the node, ranks the nodes based on the importance of the other connected nodes of the node of interest [15]. It also highlights that if a message is generated somewhere in network, how likely is that message will reach that particular node as well. We applied PageRank algorithm to identify the importance of a node (narrator) in social network of narrators to measure the influence of a node. For community detection, We identified the communities of narrators using the methodology developed by Blondel et at [16]. Table S1 highlights the list of narrators and corresponding centrality measures and Table S2 highlights the community along with its members.

We used custom scripts written in Python and Java, NetworkX package [17] and Gephi software [18] for the visualization of the narrator network.

## SUPPLEMENTARY FLES

**Table S1:** List of narrators from Sahih Bukhari and different centrality measurements for each narrator.

**Table S2:** List of narrators from Sahih Bukhari and their community.

All Supplementary files are stored for reviewers in: https://github.com/tanviralambd/Sahih_Bukhari_SNA.

## ACKNOWLEDGMENT

None.